\pdfoutput=1

\documentclass[11pt]{article}

\usepackage[]{acl}

\usepackage{times}
\usepackage{latexsym}
\usepackage{multirow}
\usepackage{comment}
\usepackage{bm}
\usepackage{enumerate}
\usepackage{amsmath,amsfonts}
\usepackage{enumitem}
\usepackage{natbib}

\usepackage[T1]{fontenc}

\usepackage[utf8]{inputenc}

\usepackage{microtype}

\usepackage{graphicx} 
\usepackage{amsmath}
\usepackage{makecell}

\newcommand{\xhdr}[1]{{\noindent\bfseries #1}.}
\newcommand{\xhdrnd}[1]{{\noindent\bfseries #1}} 

\newcommand{\peng}[1]{{{\color{purple!60!blue}{[peng: #1]}}}}

\newcommand{\chao}[1]{{{\textcolor{teal}{[chao: #1]}}}}

\def\wg{\textsuperscript{1}}
\def\ws{\textsuperscript{2}}

\title{Improving Time Sensitivity for Question Answering over Temporal Knowledge Graphs}

\author{
Chao Shang\wg, Guangtao Wang\wg, Peng Qi\wg, 
Jing Huang\ws
\thanks{ \hspace{1mm} Work done at JD AI Research.} \\
  \wg JD AI Research\\
  \ws Alexa AI, Amazon\\
  \texttt{\{chao.shang3, guangtao.wang, peng.qi\}@jd.com} \\ \texttt{ jhuangz@amazon.com}\\}

\begin{document}
\maketitle
\begin{abstract}

Question answering over temporal knowledge graphs (KGs) efficiently uses facts contained in a temporal KG, which records entity relations and when they occur in time, to answer natural language questions (e.g., ``Who was the president of the US before Obama?''). These questions often involve three time-related challenges that previous work fail to adequately address: 1) questions often do not specify exact timestamps of interest (e.g., ``Obama'' instead of 2000); 2) subtle lexical differences in time relations (e.g., ``before'' vs ``after''); 3) off-the-shelf temporal KG embeddings that previous work builds on ignore the temporal order of timestamps, which is crucial for answering temporal-order related questions. In this paper, we propose a time-sensitive question answering (TSQA) framework to tackle these problems. TSQA features a timestamp estimation module to infer the unwritten timestamp from the question. We also employ a time-sensitive KG encoder to inject ordering information into the temporal KG embeddings that TSQA is based on. With the help of techniques to reduce the search space for potential answers, TSQA significantly outperforms the previous state of the art on a new benchmark for question answering over temporal KGs, especially achieving a 32\% (absolute) error reduction on complex questions that require multiple steps of reasoning over facts in the temporal KG.


\end{abstract}

\section{Introduction}

\begin{figure}[ht]
  \centering
  \includegraphics[width=0.99\linewidth]{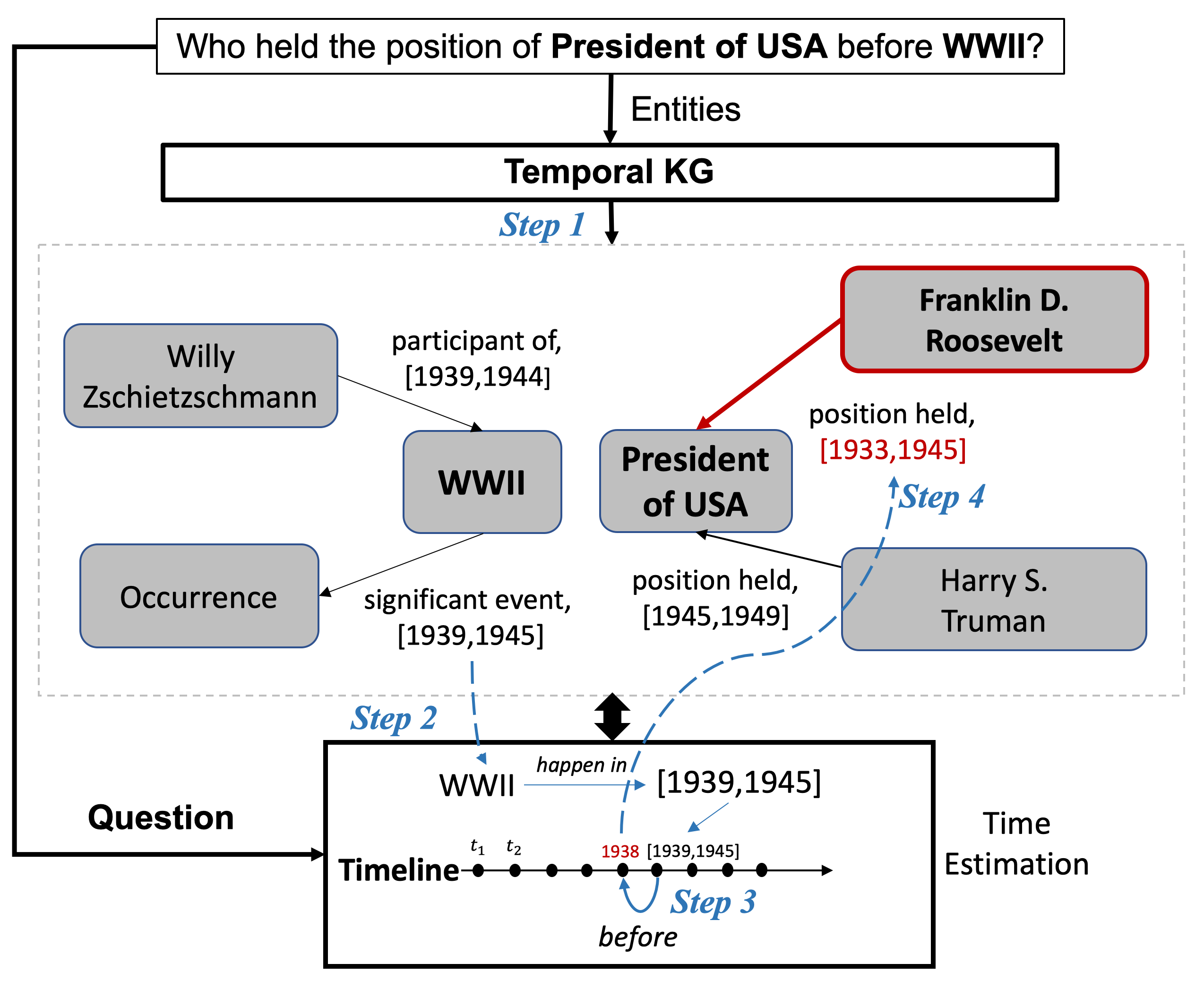}
  \caption{An example of complex temporal question on a temporal KG. }
  \label{fig:example}
\end{figure}


Temporal knowledge graphs (KGs) record the relations between entities and the timestamp or time period when such relation hold, e.g., in the form of a quadruple: (Franklin D. Roosevelt, position held, President of USA, [1933, 1945]).
This makes them a perfect source of knowledge to answer questions that involve knowledge of when certain events occurred as well as how they are related temporally (see Figure \ref{fig:example} for an example).
Unlike question answering (QA) over non-temporal KGs that is mainly concerned with relational inference, a core challenge in temporal KGQA is correctly identifying the time of reference mentioned explicitly or implicitly in the question, and locating relevant facts by jointly reasoning over relations and timestamps.

Inspired by work on relational KGQA \citep{huang2019knowledge, saxena2020improving}, where knowledge graph embeddings \cite{dasgupta2018hyte, garcia2018learning, goel2020diachronic, wu2020temp, lacroix2020tensor} learned independently of question answering are used as input to KGQA models, previous work \cite{saxena2021question} employs temporal KG embeddings to attack the problem of temporal KGQA.
Despite its relative success on simple temporal questions that directly queries facts in the KG with one out of the four facts left as the answer (e.g., ``When was Franklin D. Roosevelt the President of USA?'' or ``What position did Franklin D. Roosevelt hold between 1933 and 1945?''), this approach still struggles to handle questions that require multiple steps of relational-temporal reasoning (e.g., the example in Figure \ref{fig:example}).
 
We identify three main challenges that hinder further progress on temporal KGQA.
Firstly, complex temporal questions often require inferring the correct point of reference in time, which is not considered by previous work.
For instance, to correctly answer the question in Figure \ref{fig:example}, it is crucial that we first identify that World War II took place between 1939 and 1945, and look for entities with the desired relation with President of USA in the time interval specified by these times.
Secondly, unlike entity relations, which are usually expressed in natural language with a handful of content words that correspond well with their recorded relations in KGs (e.g., ``What position did ... hold ...'' vs the ``position held'' relation), temporal relations often involve just one or two prepositions (e.g., ``before'' or ``during'') and are expressed only implicitly in temporal KGs (e.g., nowhere is it clearly stated that 1931 is earlier than, or before, 1934, by a gap of 3 years).
As a result, a small lexical change can drastically alter the temporal relation expressed by the question, and therefore the answer set.
Thirdly, previous work on temporal KGQA build on temporal KG embeddings, where each timestamp is assigned a randomly initialized vector representation that is jointly optimized with entity and relation representations to reconstruct quadruples in the KG from embeddings.
While sound as a standalone method for encoding knowledge in temporal KGs, this approach does not guarantee that the learned timestamp representations can recover implicit temporal relations like temporal orders or distance, which are crucial for temporal KGQA.

In this paper, we propose a time-sensitive question answering framework (TSQA) to address these challenges.
We first equip the temporal KGQA model with a time estimation module that infers the unstated timestamps from questions as the first step of reasoning, and feed the result into relational inference as a reference timestamp.
Even without explicit training data for this module, the explicit factorization of the problem yields significant improvement over previous work on complex questions that require reasoning over multiple temporal quadruples.
To improve the sensitivity of our question encoder to time relation words, we also propose auxiliary contrastive losses that contrast the answer prediction and time estimation for questions that differ only by the time relation word (e.g., ``before'' vs ``after'').
By leveraging the mutual exclusiveness of answers and the prior knowledge regarding potential time estimates from different time relation words, we observe further improvements in model performance on complex questions.
Next, to learn temporal KG embeddings with prior knowledge of temporal order and distance built in, we introduce an auxiliary loss of time-order classification between each pair of timestamp embeddings.
As a result, the knowledge in the temporal KG can be distilled into the entity, relation, and timestamp embeddings where the timestamp embeddings can naturally recover order and distance information between the underlying timestamps, thus improving the performance of temporal KGQA where such information is crucial.
Finally, we enhance TSQA with KG-based approaches to narrow the search space to speed up model training and inference, as well as reduce the number of false positives in model prediction.
As a result, TSQA outperforms the previous state of the art on the \textsc{CronQuestions} benchmark \citep{saxena2021question} by a large margin.

To summarize, our contributions in this paper are: a) we propose a time-sensitive question answering framework (TSQA) that performs time estimation for complex temporal answers; b) we present contrastive losses that improve model sensitivity to time relation words in the question; c) we propose a time-sensitive temporal KG embedding approach that benefits temporal KGQA; d) with the help of KG-based pruning technique, our TSQA model outperforms the previous state of the art by a large margin.

\section{Related Work}

\xhdr{Temporal Knowledge Graph Embedding}
Knowledge graph embedding learning \cite{bordes2013translating,yang2014embedding, trouillon2016complex, dettmers2018convolutional,shang2019end,sun2019rotate,tang2019orthogonal,ji2021survey} has been an active research area with applications directly in knowledge base completion and relation extractions.
Recently, there are several works that extended the static KG embedding models to temporal KGs. \citet{jiang2016towards} first attempt to extend TransE \cite{bordes2013translating} by adding a timestamp embedding into the score function. Later, Hyte \cite{dasgupta2018hyte} projects each timestamp with a corresponding hyperplane and utilizes the TransE score in each space. \citet{garcia2018learning} extend TransE and DistMult by utilizing recurrent neural networks to learn time-aware representations of relation types.
TCompLEx \cite{lacroix2020tensor} extends the ComplEx with time based on the canonical decomposition of tensors of order 4.

\xhdr{Temporal QA on Knowledge Graph}
Temporal QA have mostly been studied
in the context of reading comprehension. 
ForecastQA \cite{jin2021forecastqa} formulates the forecasting problem as a multiple-choice question answering task, where both the articles and questions include the timestamps. The recent released TORQUE \cite{ning2020torque} is a dataset that explores the temporal ordering relations between events described in a passage of text. 

Another direction is the temporal question answering over knowledge bases (KB)~\cite{jia2018tequila,jia2018tempquestions}, which retrieves time information from the KB. TempQuestions \cite{jia2018tempquestions} is a KGQA dataset specifically aimed at temporal QA. 
Based on this dataset, \citet{jia2018tequila} design a method that decomposes and rewrites each question into nontemporal sub-question and temporal sub-question. Here the KG used in TempQuestions is based on a subset of FreeBase which is not a temporal KG. Later \citet{jia2021complex} proposes a first end-to-end system (EXAQT) for answering complex temporal questions, which takes advantage of the question-relevant compact subgraphs within the KG, and relational graph convolutional networks \cite{schlichtkrull2018modeling} for predicting the answers. All previous datasets only include a limited number of temporal questions. 
Recently, a much larger 
temporal KGQA dataset \textsc{CronQuestions} \cite{saxena2021question} is released, which includes both the temporal questions and the temporal KG with time annotation for all edges.
Based on this dataset, the CronKGQA model \cite{saxena2021question} is presented that exploits recent advances in Temporal KG embeddings and achieves performance superior to all baselines. 

\begin{figure*}[ht]
  \centering
  \includegraphics[width=\textwidth]{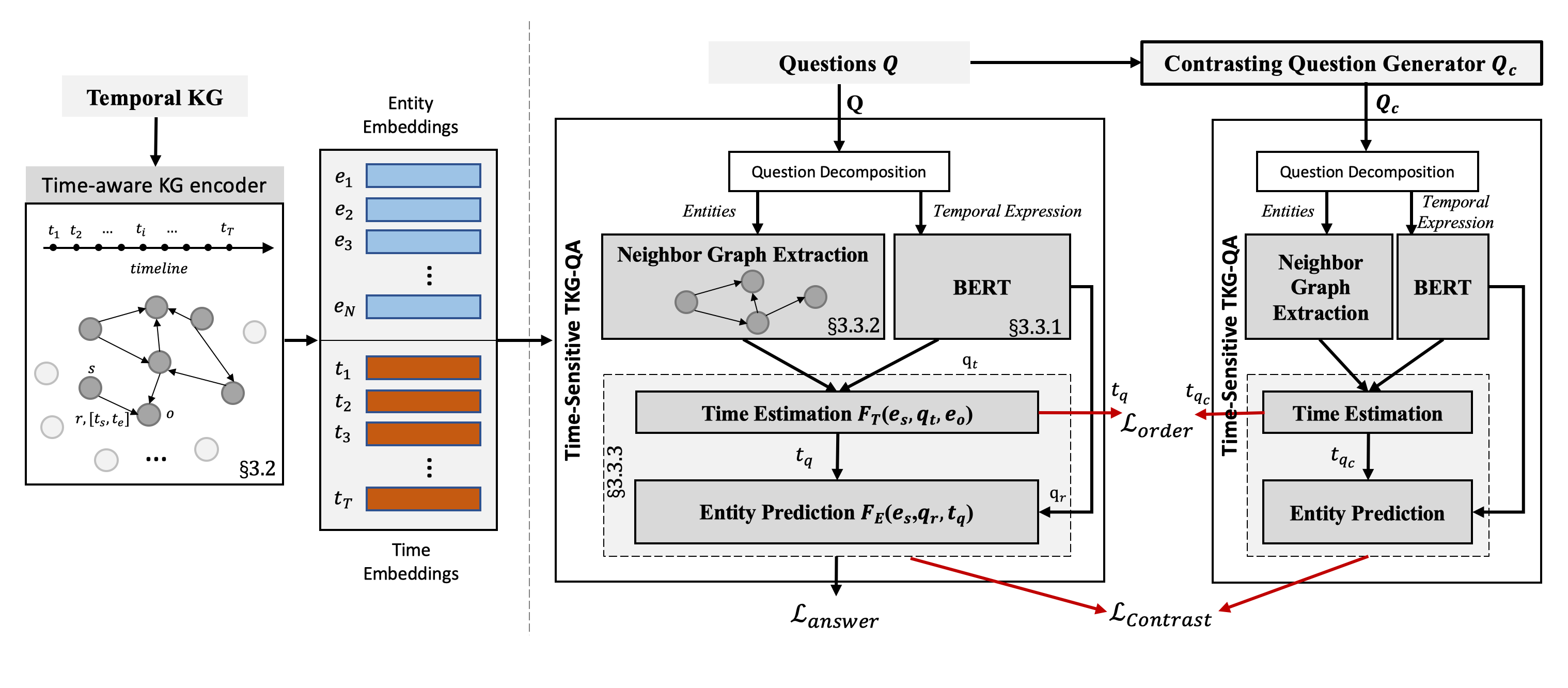}
  \caption{The architecture of our TSQA model (\textbf{Left}: Time-aware TKG encoder; \textbf{Right}: Time-Sensitive TKG-QA). 
  }
  \label{fig:architecture}
\end{figure*}


\section{Method}


In this section, we first give the problem definition of temporal question answering over temporal knowledge graph. Then, we introduce the framework to solve this problem, which 
integrates time sensitivity into KG embedding and answer inference. Finally, we describe the key modules of our proposed system in details.  


\subsection{Problem Definition and Framework}

\xhdrnd{QA on Temporal KG} aims at finding out the answer from a given temporal KG $G = (\mathcal{V}, \mathcal{E}, \mathcal{R}, T)$ for a given free-text temporal question $Q$ containing implicit temporal expression, and the answer is either an entity of entity set $\mathcal{V}$ or a timestamp of timestamp set $T$. Here, $\mathcal{E} \subseteq \mathcal{V} \times \mathcal{V}$ is a set of edges, and $\mathcal{R}$ is the set of relations. 
Edge from a quadruple $(s, r, [t_s, t_e], o)$ indicates the relation $r\in \mathcal{R}$ holds between subject entity $s$ and object entity $o$ during time interval $[t_s, t_e]$ ($t_s < t_e$ and $t_{e/s}\in T$).






\xhdr{Framework} Our framework resorts to KG embeddings along with pretrained language models to perform temporal KGQA.  Figure~\ref{fig:architecture} shows the architecture which consists of two modules: 1) time-aware TKG encoder; 2) time-sensitive question answer. 

The time-aware TKG encoder extends the existing TKG embedding method by adding an auxiliary time-order learning task to consider the quadruple orders. And the time sensitive QA module first performs neighboring graph extraction to reduce the search space for question answer, then performs joint training for answer/time prediction and time-sensitive contrastive learning to enhance the model ability in capturing temporal signals in free-text question. Next, we will introduce these two modules in details.

\subsection{Time-aware KG Encoder}

We first briefly review a time-aware KG embedding method based on TCompLEx \cite{lacroix2020tensor} since it has been used in \cite{saxena2021question} for TKGQA and shows competitive performance. Next, we show that how to perform TCompLEx on temporal KG, then analyze its weakness in TKGQA especially for complex question and further overcome such weakness by introducing an auxiliary time-order learning task in TKG embedding.


\xhdr{TCompLEx for TKG} TCompLEx is an extension of ComplEx considering time information, which not only encodes the entity and relation to complex vectors, but also maps each timestamp to a complex vector. To perform TCompLEx over temporal KG in our problem definition, we first reformulate each quadruple to a set of new quadruples by
\begin{equation}
    (s, r, [t_s, t_e], o) = \{(s, r, t, o)| t_s \leq t \leq t_e\}
\end{equation}

Let $\bm{e}_s$, $\bm{e}_r$, $\bm{e}_t$, $\bm{e_o}\in \mathbb{C}^{d}$ be the complex-value embeddings of $s, r, t, o$, respectively. Then, TCompLEx scores each quadruple $(s, r, t, o)$ by
\begin{equation}
\mathcal{S}(s,r,o,t) = Re(\langle \bm{e}_s, \bm{e}_r, \bm{e}_o, \bm{e}_t \rangle)
\end{equation}
where Re(.) denotes the real part of a complex vector, and $\langle\rangle$ denotes the multi-linear product.

Finally, we use a loss function similar to the negative sampling loss for effectively TCompLEx training.
\begin{equation}\label{eq:tcomplex_loss}
\begin{split}
   L_{TC} = -log(\phi(\gamma - \mathcal{S}(s,r,o,t))) \\ - \frac{1}{K}\sum_{i=1}^{K}(log(\phi(\mathcal{S}(s'_{i},r,o'_{i},t'_{i}) - \gamma))),
\end{split}
\end{equation}
where $\gamma$ is a fixed margin, $\phi$ is the sigmoid function, $(s'_{i},r,o'_{i},t'_{i})$ is the $i$-th negative quadruple.

According to the loss function in equation~\ref{eq:tcomplex_loss}, we observe that TCompLEx only cares about whether the quadruple is true or false and ignores the orders of different quadruples occur. However, the time orders are critical to find the correct answer in knowledge graphs. For example, to answer the `Who is the President of USA before William J. Clinton?'', we need not only the two facts (President of USA, Position Held, Ronald Reagan, [1981, 1989]) and (President of USA, Position Held, William J. Clinton, [1993, 2001]), but also the time order of these facts. To overcome such a limit of TCompLEx in TKGQA, we introduce an auxiliary time-order learning task over time-embeddings.


\xhdr{Time-order learning in TKG} To keep the time order in embedding spaces, we first sort the timestamps in $T$ by an ascending order and get $(t_1, t_2, \cdots, t_{|T|})$ and $t_i < t_j$ if $1\leq i < j \leq |T|$. Let $\bm{t}_i =  [Re(\bm{e}_{t_i}), Im(\bm{e}_{t_i})] \in \mathbb{R}^{2d}$ be the concatenation the real and imaginary components of embedding $\bm{e}_{t_i}$ of timestamp $t_i$. Inspired by position embedding in \cite{vaswani2017attention}, we first initialize the timestamp embedding $\bm{t}_i$ as follows.
\begin{equation}
\begin{aligned}
\bm{t}_{i}[2k] = \sin(\frac{i}{10000^{2k/2d}})\\
\bm{t}_{i}[2k + 1] = \cos(\frac{i}{10000^{2k/2d}})
\end{aligned}
\end{equation}
where $0\leq k \leq d-1$.

Afterwards, for any pair of timestamps ($t_i, t_j$),  we calculate the probability of time order as:
\begin{equation}
    p_{t}(i,j) = sigmoid((\bm{t}_1-\bm{t}_2)^T \bm{W}_{t}),
\end{equation}
where $\bm{W}_t \in \mathbb{R}^{2d}$ represents a parameter vector.

Based on the time-order probabilities, we introduce a binary cross-entropy loss as a time-order constraint over timestamp embeddings as follow:
\begin{equation}
\begin{aligned}
    L_{TO} =& -\delta(i, j)\log(p_{t}(i, j)) \\ &- (1 -\delta(i, j))\log(1 - p_{t}(i, j)),
    \end{aligned}
\end{equation}
where $\delta(i,j) = 1$ if $t_i < t_j$ else $\delta(i,j) = 0$.

\xhdr{Joint-training} 
A weighted sum of T-CompLEx training loss and time-order constraint is considered as the final objective function for the joint training for TKG embedding.
\subsection{Time-Sensitive TKG-QA}

In this section, we introduce our time-sensitive question answering module from the following aspects in details: 1) question decomposition which divides the questions as entities and relations described in free-text; 2) entity neighboring sub-graph extraction which reduces the search space of candidate timestamps and answer entities; and 3) time-sensitive question answer which explores the time information implied in both KG and questions to help the model find the answer.

\subsubsection{Question Decomposition and Encoder}
\label{sec3_5}

For each question $Q$, we first identify all the entities \{Ent$_1$, Ent$_2$, $\cdots$, Ent$_k$\} in $Q$ which also appear in KG $G$, i.e., Ent$_i \in E$ ($1\leq i \leq k$).
Then, by replacing the entities in question $Q$ with special token [\textit{subject}] and [\textit{object}] in order, we obtain an entity-independent temporal relation description in free-text named \textbf{temporal expression} $\hat{Q}$.

Taking the question ``When did \textbf{Obama} hold the position of \textbf{President of USA}?'' as an example, by replacing the identified entities ``President of USA'' and ``Obama'', we get its temporal expression as ``When did \textbf{\textit{subject}} hold the position of \textbf{\textit{object}}?''.


Next [CLS] + $\hat{Q}$ are fed into BERT that outputs [CLS] token embedding as $\bm{e}_{q}\in \mathbb{R}^{d_{bert}}$, where $d_{bert}$ is the output dimension of BERT, and two kinds of question representations as follows.
\begin{align}
    \bm{q}_r = \bm{W}_{q}^{r}(\delta(\bm{W}\bm{e}_{q}))\\
    \bm{q}_t = \bm{W}_{q}^{t}(\delta(\bm{W}\bm{e}_{q})),
\end{align}
where $\bm{q}_r,  \bm{q}_t \in \mathbb{R}^{2d}$ represents the embedding of relation and time implied in question, respectively. $\bm{W}\in {\mathbb{R}^{d_{bert}\times 2d}}$, $\bm{W}_{q}^{r}$, $\bm{W}_{q}^{t} \in \mathbb{R}^{ 2d \times 2d}$ are the parameter matrix, and $\delta$ represents the activation function. Finally, to facilitate the calculation with KG embeddings, we reformulate $\bm{q}_r, \bm{q}_t$ in complex space as:
\begin{align}
    \bm{q}_r = \bm{q}_r[0:d] + \sqrt{-1} \cdot \bm{q}_r[d:2d]\\
    \bm{q}_t = \bm{q}_t[0:d] + \sqrt{-1} \cdot \bm{q}_t[d:2d]
\end{align}



\subsubsection{Entity Neighbor Graph Extraction}

Let \{Ent$_1$, Ent$_2$, $\cdots$, Ent$_k$\} be the $k$ entities extracted from question $Q$, we first extract the $m$-hop neighboring sub-graph $G_i$ for each entity Ent$_i$. Then, by combining these $k$ sub-graphs, we obtain the search graph $G_q$ for question $Q$: $G_q =  	\cup_{i=1}^{k}G_i$. Suppose that $E_q$ and $T_q$ are the sets of entities and timestamps appearing in $G_{q}$, respectively, they constitute the search space of time and entity prediction in our TKG-QA method. In training stage, we set the hop number $m$ as the minimum value which results in correct answer entity appearing in $G_q$. In testing stage, we set $m$ as the largest hop number used in training stage. In practice, the size of graph $G_q$ in usually much smaller than that of whole graph $G$. For example, in CronKGQA, 
the average value of $|G_q|/|G|$ is about 3\%.

Entity Neighboring graph extraction aims at reducing the search space of candidate timestamps and answer entities.
This results in not only more efficient training procedure, but also performance improvement of question answer because a larger number of candidates usually means a much more difficult learning problem. 
\subsubsection{Time-Sensitive Question Answering}
\label{tbreasoning}




For temporal question answer over KG, the interaction of time and answer entity prediction is very important since the time range brings a strong constraint on the search space of answers. However, the existing method \cite{saxena2021question} usually performs such two predictions independently which results in poor performance especially for complex questions which need to consider multiple facts to get the answer. To overcome this limitation, we directly feed the intermediate time representation $t_q$ learned from time estimation to answer prediction to enhance the interaction of these two tasks.

\noindent \textbf{Time Estimation.}
Based on the embeddings $\bm{e}_s$ and $\bm{e}_o$ of subject entity $s$ and object entity $o$ from KG and the time embedding $\bm{q}_{t}$ from a question, we design the time estimation function $F_T$ for learning the time embedding $\bm{t}_q$ as follows:\footnote{A simple temporal question might contain the timestamp (e.g. 2001). In this case, we set $\bm{t}_q$ as the linear combination of this learned time embedding and the timestamp embedding from KG.} 
\begin{equation}
\begin{aligned}
\bm{t}_q &=  F_T(\bm{e}_s, \bm{q}_{t}, \bm{e}_o) \\
 & =  \bm{W}_{q}^{t}([Re(\langle \bm{e}_s, \bm{q}_{t}, \bm{e}_o\rangle), Im(\langle \bm{e}_s, \bm{q}_{t}, \bm{e}_o\rangle)]),
\end{aligned}
\label{eqa_time_emb}
\end{equation}
where $\bm{W}_{q}^{t} \in \mathbb{R}^{2d\times 2d}$ represents the parameter matrix. [.] is the concatenation function, $Re(.)$ denotes the real part of a complex vector and $Im(.)$ is the imaginary part.

After getting the time embedding w.r.t. question $\bm{t}_q$, for timestamp prediction, the following score function to estimate the score for each timestamp $t \in T_q$ as follow:
\begin{equation}
\mathcal{S}_t = Re(\langle \bm{t}_q, \bm{t} \rangle)
\end{equation}

\xhdr{Entity Prediction} In enhance the interaction between time prediction and answer prediction, we update the embedding of entity w.r.t. question by considering time embedding $\bm{t_{q}}$ by an entity function $F_E$ as follow:
\begin{equation}
\bm{e}_q = F_E(\bm{e}_s, \bm{q}_r, \bm{t}_q) = \langle \bm{e}_s, \bm{q}_r, \bm{t}_q \rangle
\end{equation}
Finally, we score the entity $e\in E_{q}$ by:
\begin{equation}
\mathcal{S}_e = Re(\langle \bm{e}_q, \bm{e} \rangle)
\end{equation}


The answer entity of the question is either timestamp or entity. Let $\mathcal{S}_a$ be the answer score and thus $\mathcal{S}_a = \mathcal{S}_t$ or $\mathcal{S}_e$ when the answer is timestamp or entity. Suppose $C$ represents the number of candidate answers (i.e., $C$ = $|E_{q}| + |T_{q}|$), then we can define the probability of $i$-th candidate answer being true as:
\begin{equation}
    P_{a,i} = \frac{\exp(\mathcal{S}_{a,i})}{\sum_{j=1}^{C}\exp(\mathcal{S}_{a,j}))}.
\end{equation}
Finally, we train the answer prediction model by minimizing the cross-entropy loss as follow:
\begin{equation}
    L_{answer}  = -\sum_{i}^{C} \bm{y}_{i} \log(P_{a,i}),
\end{equation}
where $\bm{y}_i$ = 1 if the $i$-th candidate is the true answer, otherwise $\bm{y}_i$ = 0.

\subsubsection{Temporal Contrastive Learning}

The temporal question answer system should be sensitive to the temporal relation implied in the question. For example, the answer of ``What does happen before a given event?'' is quite different from that of ``What does happen after a given event?''. Existing works on TKG-QA usually resort to pre-trained language models for question understanding. But these models are not sensitive to the difference of temporal expressions in free-text~\cite{ning2020torque,dhingra2021time,shang2021open,han2021econet}, and thus prone to wrong predictions. 

To make the system 
sensitive to the temporal relation implied in question, we resort to a contrastive learning method: 
we construct a contrastive question to the original question, then add auxiliary contrastive learning tasks to distinguish the latent temporal representation and prediction results coming from the pair of contrastive questions.


\xhdr{Contrastive Question Generation} To generate the contrastive question $\bar{Q}$ for the given question $Q$, we first extract all the temporal words based on large number of questions in temporal question answer dataset, and then build a contrastive word pair dictionary by finding the antonyms. The dictionary consists of $\mathcal{D}_{contr}$ = \{(first, last), (before, after), (before, during), (during, after), (before, when), (when, after)\}. Based on such dictionary, we replace the temporal word in given question $Q$ by its antonym to generate its contrastive question $\bar{Q}$.




\xhdr{Contrastive time order learning} For the contrastive question pair $Q$ and $\bar{Q}$, we follow the same encoder in Eq. \ref{eqa_time_emb} to get the corresponding time-aware embeddings $\bm{t}_q$ and $\bm{t}_{q_c}$, respectively. Meanwhile, according to the contrastive temporal word pair dictionary, suppose that we pickup the pair (word$_1$, word$_2$) $\in \mathcal{D}_{contr}$ for contrastive question construction, we can construct a question order label $y_o$: $y_o$ = 0 if $\bar{Q}$ is achieved by replacing word$_1$ as word$_2$, else $y_o$ = 1.

Afterward, we distinguish the temporal orders implied by word$_1$ and word$_2$ by predicting of the order label $y_o$ based on $\bm{t}_q$ and $\bm{t}_{q_c}$ as follow:
\begin{align}
    p_o & = sigmoid((\bm{t}_{q} - \bm{t}_{q_c})^{T}\bm{W}_o)\\
    \mathcal{L}_{\text{order}} & = -{{y}_o \log(p_{o})} - {(1-{y}_o) \log(1-p_{o})},
\end{align}
where $\bm{W}_o \in \mathbb{R}^{2d}$ represents the parameter vector to be learned.



\xhdr{Answer-guided Contrastive Learning} Let $\bm{S} = [\bm{s}_{1}, \cdots, \bm{s}_{C}]$, $\bm{\bar{S}} = [\bm{s}_{1}, \cdots, \bm{s}_{C}]$ be the answer scores w.r.t. questions $Q$ and its contrastive question $\bar{Q}$, respectively, where $C = |E_q| + |T_q|$. By stacking these two scores together, we get $\bm{S}_{q}$ = $[\bm{{S}}; \bm{\bar{S}}]\in \mathbb{R}^{2\times C}$. Then, we can apply softmax over $\bm{S}_{q}$ along the last dimension and get the probability scores $\bm{P}_{q} =softmax(\bm{S}_{q})\in \mathbb{R}^{2\times C}$ and $sum(\bm{P}_{q}[:,i]) = 1$ for $i= 1, \cdots, C$.


Due to the fact that the answers of question ${Q}$ are definitely not for question $\bar{Q}$, we construct an answer-guided learning labels as $\bm{y}_{a} = [y_1,\cdots, y_C]$, where $y_i$ = 1 if and only if the $i$-th candidate is true answer for ${Q}$, otherwise $\bm{y}_i$ = 0. Then, we get an answer-guided contrastive loss as follow:
\begin{equation}
    \mathcal{L}_{\text{contrast}} = -{\frac{1}{C}}\sum_{j=0}^{C}{y}_i\log(\bm{P}_{q}[0,i])
\end{equation}



\xhdr{Joint Training} We combine the answer prediction loss and contrastive losses as the final objective function for joint training:
\begin{equation}
    Loss = \mathcal{L}_{answer} + \lambda_{o} \cdot \mathcal{L}_{order} + \lambda_{c} \cdot \mathcal{L}_{contrast},
\end{equation}
where $\lambda_{o} > 0$, $\lambda_{c} >0$ are the weight factors to make tradeoffs between different losses.

\section{Experiments}
In this section, we conduct experiments to assess the effectiveness of our proposed method TSQA for TKG-QA. 
Our experimental results show that our approach obtains significant improvements over the baseline models.

\begin{table}[!ht]
    \centering 
    \small
    \begin{tabular}{c|c|c|c}
    \hline
    Category & Train & Dev &  Test \\ \hline \hline
    Simple Entity & 90,651 & 7,745 & 7,812 \\ 
    Simple Time & 61,471 & 5,197 & 5,046 \\ 
    Before/After & 23,869 & 1,982 & 2,151\\ 
    First/Last& 118,556& 11,198& 11,159\\ 
    Time Join& 55,453& 3,878& 3,832\\ 
    \hline  
    Entity Answer & 225,672&  19,362&  19,524\\ 
    Time Answer&  124,328&  10,638&  10,476\\ 
    \hline 
    Total&  350,000&  30,000&  30,000\\ 
    \hline 
    \end{tabular}
    \caption{\textsc{CronQuestions} dataset statistics as well as the numbers of questions across different types of reasoning required and answer types.}
    \label{DatasetStats}
\end{table}

\subsection{Experimental Setup}

\xhdr{Data} \textsc{CronQuestions}\footnote{\url{https://github.com/apoorvumang/CronKGQA}} is the largest known Temporal KGQA dataset consisting of two parts: a KG with temporal annotations, and a set of free-text questions requiring temporal reasoning. This Temporal KG has 125k entities and 328k facts (quadruples), while a set of 410k questions is given. The facts have the time spans in the edge. These time spans or timestamps were discretized to years.


This dataset consists of questions that can be categorized into two groups based on their answer type: entity questions where the answer is an entity in the KG, and time questions where the answer is a timestamp.
The authors also categorize these questions into ``simple reasoning'' (including \textit{simple entity} and \textit{simple time} subtypes) and ``complex reasoning'' (including \textit{before/after}, \textit{first/last} and \textit{time join} subtypes). 
Table \ref{DatasetStats} provides the number of questions across different categories. Complex questions require complex temporal reasoning which takes advantage of multiple facts and temporal order of these facts.

\begin{table*}[!ht]
    \centering
    \small
    \begin{tabular}{c||c|c|c|c|c||c|c|c|c|c}
      \hline
      \multirow{3}{4em}{Model}  &  \multicolumn{5}{c||}{Hits@1} &  \multicolumn{5}{c}{Hits@10}  \\ 
      \cline{2-11}
       &  & \multicolumn{2}{c|}{Question Type}& \multicolumn{2}{c||}{Answer Type} &  & \multicolumn{2}{c|}{Question Type}& \multicolumn{2}{c}{Answer Type}\\
      \cline{2-11}
       & Overall & Complex & Simple & Entity & Time & Overall & Complex & Simple & Entity & Time  \\ \hline 
      EmbedKGQA & 0.288 & 0.286 & 0.290 & 0.411 & 0.057 & 0.672 & 0.632 & 0.725 & 0.850 & 0.341 \\
      T-EaE-add & 0.278 &0.257& 0.306& 0.313& 0.213& 0.663& 0.614& 0.729& 0.662& 0.665 \\
      T-EaE-replace & 0.288 &0.257& 0.329& 0.318& 0.231& 0.678& 0.623& 0.753& 0.668& 0.698 \\
      CronKGQA & 0.647 & 0.392 & 0.987 & 0.699 & 0.549 & 0.884 & 0.802 & 0.992 & 0.898 & 0.857 \\
      \hline
      TSQA & \textbf{0.831} & \textbf{0.713} & \textbf{0.987} & \textbf{0.829} & \textbf{0.836} & \textbf{0.980} & \textbf{0.968} & \textbf{0.997} & \textbf{0.981} & \textbf{0.978} \\
      \hline
    \end{tabular} 
    \caption{Comparison of different TKG-QA models on \textsc{CronQuestions} dataset.}
    \label{ResultsTable2}
\end{table*}

\smallskip
\xhdrnd{Evaluation Metrics} include Hits@1 and Hits@10 , which is the standard evaluation metrics on \textsc{CronQuestions}~\cite{saxena2021question}.

\smallskip
\xhdr{Hyper-parameter setting} We train the TSQA models by setting the hyper-parameters as: learning rate = $\{$1$e^{-4}$, 2$e^{-5}$, 1$e^{-5}$ $\}$, $\lambda_{o}$ = $\{$0.5, 1.0, 2.0, 3.0, 5.0$\}$ and $\lambda_{c}$ = $\{$0.5, 1.0, 2.0, 3.0, 5.0$\}$, and pick up the best hyper-parameters on dev set 
by the overall Hits@1 metrics. 
Our models are implemented by PyTorch and trained using NVIDIA Tesla V100 GPUs.

\smallskip
\xhdr{Baselines} We select several recent SOTA TKG-QA models as our baselines as follow:
\begin{itemize}[leftmargin=*]
    \item  EmbedKGQA \cite{saxena2020improving} is the first method to use KG embeddings for the multi-hop KGQA task. It uses ComplEx~\cite{trouillon2016complex} embeddings and can only deal with non-temporal KGs and single entity questions.

    \item T-EaE-add/replacement~\cite{saxena2021question} are two modifications of KG enhanced language model EaE~\cite{fevry2020entities}, which integrates entity knowledge into a transformer-based language model and has been used for TKG-QA~\cite{saxena2020improving}. T-EaE-add has all grounded entities and time spans marked in the question, and T-EaE-replace replaces the BERT embeddings with the entity/time embeddings instead of adding them with token embeddings.
    
    \item CronKGQA \cite{saxena2021question} extends EmbedKGQA to the temporal QA task, and takes advantage of the temporal KG embeddings to answering temporal questions. This is the current SOTA model on \textsc{CronQuestions}.
\end{itemize}




\subsection{Main Results}
Table \ref{ResultsTable2} compares different TKG-QA methods in terms of Hits@1 and Hits@10. From this table, we observe that: 1) our proposed TSQA has achieved state-of-the-art performance in terms of all types of questions on both Hits@1 and Hits@10. 2) The performance improvement over the SOTA model is significant. TSQA outperforms the SOTA results by more than 82\% Hits@1 relative improvement (32\% absolute error reduction) on complex questions and 21\% Hits@10 relative improvement on simple questions.
These results proved the excellent performance of our proposed TSQA on question answering on the temporal knowledge graph, especially for complex temporal reasoning.

\begin{table}[!ht]
    \centering
    \small
    \resizebox{.48
\textwidth}{!}{
    \begin{tabular}{c||c|c|c|c|c}
      \hline 
      Question type & \makecell[c]{Before\\After} & \makecell[c]{First\\Last} & \makecell[c]{Time\\Join} & \makecell[c]{Simple\\Entity} & \makecell[c]{Simple\\Time}   \\ 
      \hline 
      EmbedKGQA & 0.199 &  0.324 & 0.223 & 0.421 & 0.087   \\
      T-EaE-add &0.256& 0.285& 0.175& 0.296& 0.321  \\
      T-EaE-replace& 0.256& 0.288& 0.168& 0.318& 0.346 \\
      CronKGQA & 0.288 & 0.371& 0.511& 0.988& 0.985 \\
      \hline
      TSQA & \textbf{0.504} & \textbf{0.721} & \textbf{0.799} & \textbf{0.988} & \textbf{0.987} \\
      \hline
    \end{tabular} 
    }
    \caption{Comparison of different models w.r.t. question type in terms of Hits@1.}
    \label{ResultsTable3}
\end{table}

We also compare our method with baselines in terms of Hits@1 on different subtype questions in Table \ref{ResultsTable3}. From this table, we observe that: on complex questions, our proposed TSQA model outperforms all baseline models significantly. The relative improvement is up to 75\%, 94\%, 56\%, for ``before/after'', ``first/last'' and ``Time Joint'', respectively. The first two kinds of questions are more challenging as they require a better understanding of the temporal expressions in question. Our method is better in capturing such time-sensitivity change in temporal words and thus results in great improvement. Moreover, for the simple questions, our method still keeps competitive performance compared to the SOTA model.

\subsection{Ablation Study}
To understand the contributions of the proposed modules in our method, we perform an ablation study by sequentially removing the following components from our proposed TSQA: temporal Contrastive learning (TC), time-aware TKG embeddings (TKE), entity neighboring graph extractor (NG), and time estimation for question answer (TE) in Table~\ref{AblationTable4}. It is noted that removing TKE means that we replace TKE with T-CompLEx as KG encoder, and removing NG means that we perform QA over the whole knowledge graph.

By comparing the two adjacent rows of this table, we can infer the contributions of TC, TKE, NG and TE, respectively: 1) all these modules improve the overall performance in terms of Hits@1, especially for complex questions; 2) by comparing the last two adjacent rows, the proposed time estimation brings significant Hits@1 improvement (14.5\%), since this module supplies the latent time embedding which not only enhances the interaction of timestamp estimation and answer estimation but also supplies a good anchor for finding the answer entity, which is very crucial for answering complex questions; 3) entity neighboring graph extraction gets 7.8\% Hits@1 improvement over complex questions by comparing rows ``{TC}-{TKE}'' and ``{TC}-{TKE}-{NG}'', since it significantly narrows down the search space of the candidate answers; 
4) by comparing the first three rows, time-aware TKG embedding (TKE) and temporal contrastive learning (TC) further boost the Hits@1 over complex questions. This is because the complex questions usually require the model  to capture time ordering information implied in temporal words of the question. 
And these two modules enhance temporal order learning by adding explicit time-order constraints.

\begin{table}[!ht]
    \centering
    \small
    \resizebox{.48
\textwidth}{!}{
    \begin{tabular}{l||c|c|c|c|c}
      \hline
      \multirow{3}{4em}{Model}  &  \multicolumn{5}{c}{Hits@1} \\ 
      \cline{2-6}
       &  & \multicolumn{2}{c|}{Question Type}& \multicolumn{2}{c}{Answer Type} \\
      \cline{2-6}
       & Overall & Complex & Simple & Entity & Time  \\ \hline
      TSQA & \textbf{0.831} & \textbf{0.713} & 0.987 & \textbf{0.829} & \textbf{0.836}  \\
      \hline
      -{TC} & 0.821 & 0.696 & 0.984 & 0.820 & 0.822  \\
      -{TC}-{TKE} & 0.816 & 0.688 & 0.985 & 0.816 & 0.818  \\
      -{TC}-{TKE}-{NG} & 0.757& 0.583& 0.986 & 0.797& 0.687\\
      -{TC}-{TKE}-{NG}-{TE} & 0.661 &0.412 &\textbf{0.989}& 0.719 &0.556 \\
      
      \hline
    \end{tabular} 
    }
    \caption{Results of the ablation study. ``-'' means to remove a module. 
    }
    \label{AblationTable4}
\end{table}

\section{Conclusion}
\label{sec:bibtex}

In this paper, we propose a time-sensitive question answering framework (TSQA) over temporal knowledge graphs (KGs). To facilitate the reasoning over temporal and relational facts over multiple facts, we propose a time estimation component to infer the unstated timestamp in the question. To further improve the model's sensitivity to time relation words in the question and facilitate temporal reasoning, we enhance the model with a temporal KG encoder that produces KG embeddings that can recover the implicit temporal order and distance between different timestamps, and with contrastive losses that compare temporally exclusive questions. With the help of answer search space pruning from entity neighboring sub-graphs, our TSQA model significantly improves the performance on complex temporal questions that require reasoning over multiple pieces of facts, and outperforms the previous state of the art by a large margin.



\section*{Acknowledgements}
This work is supported by the National Key Research and Development Program of China under Grant No. 2020AAA0108600.


\bibliography{anthology,custom}
\bibliographystyle{acl_natbib}




\end{document}